\documentclass[letterpaper, 10 pt, conference]{ieeeconf}

\IEEEoverridecommandlockouts               %

\usepackage[style=ieee,mincitenames=1,maxcitenames=2,maxbibnames=12,doi=false]{biblatex}

\usepackage{amsmath}
\usepackage{graphicx}
\usepackage{subcaption}
\usepackage{url}
\usepackage{csquotes}
\usepackage[american]{babel}
\usepackage{siunitx}
\usepackage{algorithm}
\usepackage{algpseudocodex}
\makeatletter
\renewcommand{\ALG@beginalgorithmic}{\footnotesize}
\makeatother

\usepackage{amssymb}

\DeclareMathOperator\erf{erf}

\usepackage{tikz}
\usepackage{hyperref}
\usepackage{cleveref}

\title{\LARGE \bf Scalable underwater assembly with reconfigurable visual fiducials}
\author{
\authorblockN{Samuel Lensgraf, Ankita Sarkar, Adithya Pediredla, Devin Balkcom, Alberto Quattrini Li}
\authorblockA{Dartmouth College}
\thanks{This project was partially supported by the NSF GRFP, CNS-1919647, 2024541, 2144624.}
}

\begin{document}
\maketitle

\begin{abstract}
    We present a scalable combined localization infrastructure deployment and task planning algorithm for underwater assembly. 
    Infrastructure is autonomously modified to suit the needs of manipulation tasks based on an uncertainty model on the infrastructure's positional accuracy. 
    Our uncertainty model can be combined with the noise characteristics from multiple devices. 
    For the task planning problem, we propose a layer-based clustering approach that completes the manipulation tasks one cluster at a time. 
    We employ movable visual fiducial markers as infrastructure and an autonomous underwater vehicle (AUV) for manipulation tasks. 
    The proposed task planning algorithm is computationally simple, and we implement it on AUV without any offline computation requirements. 
    Combined hardware experiments and simulations over large datasets show that the proposed technique is scalable to large areas. 
\end{abstract}

\section{Introduction}

Autonomous assembly of structures using drones or free-floating robots is a promising direction for creating rapidly deployable, flexibly designed structures~\cite{augugliaroFlightAssembledArchitecture2014}. 
In most real-world systems, localization relative to a reference is achieved using calibrated and fixed positional infrastructure such as motion capture systems or visual fiducials~\cite{hunt3DPrintingFlying2014, damsAerialAdditiveBuilding2020, goessensFeasibilityStudyDronebased2018, lensgrafDropletAutonomousUnderwater2021}. 
Unfortunately, these systems are not scalable as the coverage area is fixed and scaling beyond the coverage area requires redesigning the positioning technology. 

\begin{figure}[t]
    \centering
    \includegraphics[width=0.95\linewidth]{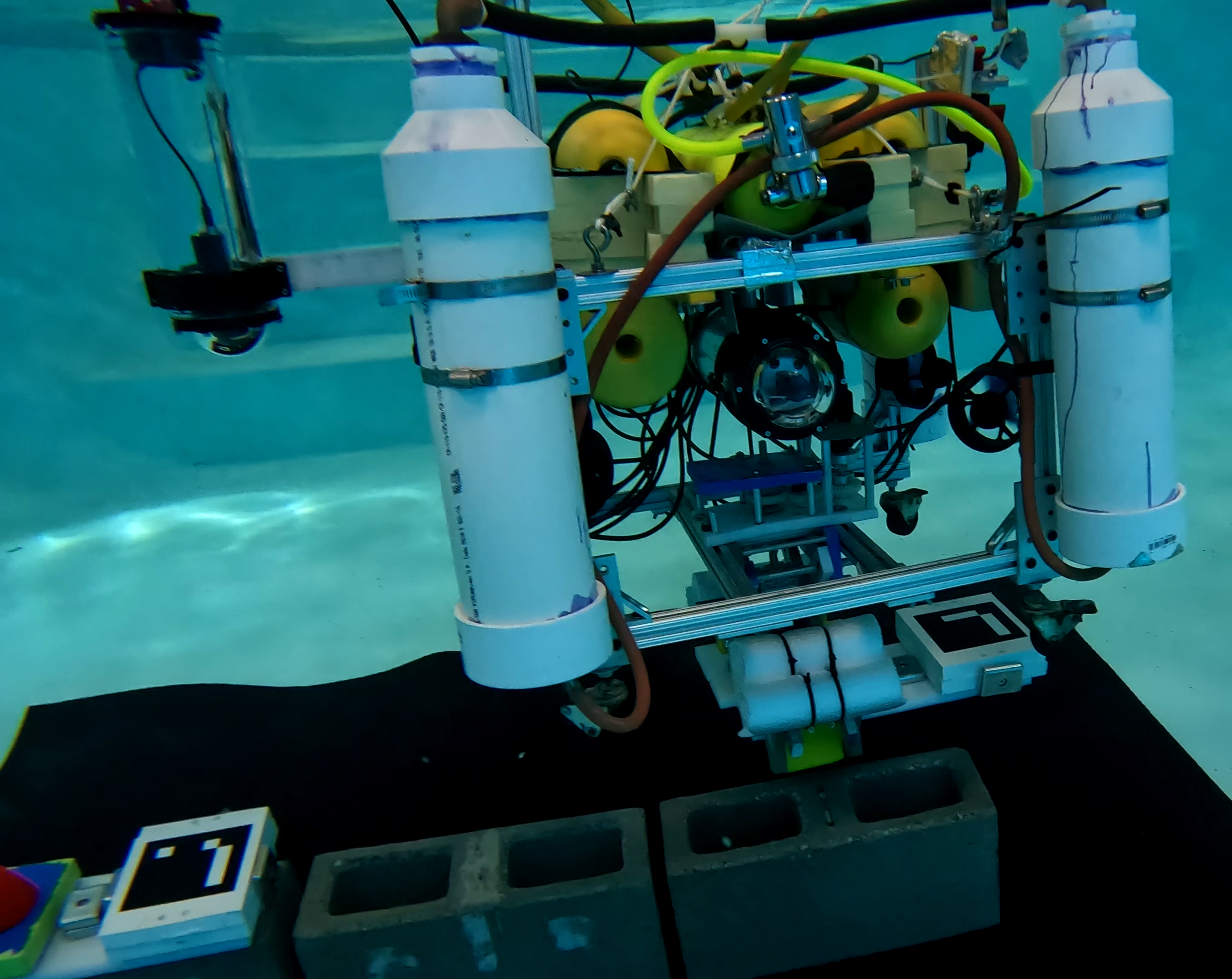}
    \caption{AUV placing a reconfigurable fiducial marker on a foundation while localizing using another marker.}
    \label{fig:hero-image}
\end{figure}

To overcome the limited coverage area of the positioning technologies, we propose to design the positioning infrastructure as a dynamic component of the construction plan. 
Our method allows localizing against large structures with minimal modification to the area around them and can also be integrated with existing underwater construction structures to make them scalable -- see Fig.~\ref{fig:hero-image} to show our robot in action while moving a fiducial marker.

Localization infrastructure is often considered to have a constant noise distribution, allowing coverage algorithms to plan based on the noise properties of the fixed infrastructure. 
However, the properties of localization infrastructure often depend on environmental factors: distance from the infrastructure, reflections, or water temperature gradients influence the accuracy of the positioning systems~\cite{paull2013auv}. 
As the accuracy depends on relative positioning between the infrastructure and the robot, high-accuracy positioning can only be provided in a small fixed area resulting in either large infrastructure requirements, or small structures.
Instead, by understanding the noise properties of the infrastructure components and modeling them accurately as a function of environmental factors, we show that it is possible to dynamically reconfigure the infrastructure to maximize the positioning accuracy of any region. 
Our noise formulation model is also conducive to sensor fusion techniques and, hence, can be extended for the case of multiple sensors. 

For a given construction structure manipulation task, we have to plan the movement of infrastructure such that the repositioned infrastructure guarantees the accuracy of the manipulation task. 
This planning is challenging and results in the combined deployment and sequencing problem.

To solve this problem, we group the manipulation tasks using a clustering algorithm that guarantees that the radius of any cluster is within the high localization accuracy achievable with the dynamic markers. 
We reposition the markers to have high accuracy for the cluster and execute the manipulation tasks, one cluster at a time. 

For implementation, we use visual fiducial markers mounted on plates that are movable on our previously developed error-correcting construction foundations~\cite{lensgrafBuoyancyEnabledAutonomous2023}. 
The AUV views these markers with a downward-facing wide-angle camera and can compute the position accuracy using our noise model. 
The implementation of our clustering-based planning algorithm is computationally simple, and we implemented it on the AUV itself without any external cloud computing requirements. 
We validated the proposed technique both experimentally and on large simulation datasets.

Our technique is particularly interesting for the case where deploying large amounts of infrastructure is impractical or expensive. 
In real-world scenarios, it is often the case as the sparse area of interest is often near a protected region that we cannot permanently mar. 
For example, in the construction of artificial reefs, the surrounding areas are critically important to protect. In hard to reach places such as caves, or very deep waters, transporting large amounts of infrastructure can be impossible.
Our technique is a foundational component towards a scalable solution for these cases. A number of practical challenges, such as  flexibility in terms of material of connected components per layer, will be subject of future studies as discussed at the end of the paper.

\section{Related work}

\textbf{Uncertainty-based planning.}
Infrastructure placement for small scale scenes is a well-studied problem. 
In particular, \textcite{magnagoEffectiveLandmarkPlacement2019} define a landmark placement algorithm that computes a static landmark placement based on certainty requirements. 
We consider the problem of dynamically altering landmark placements. 
We also develop a direct model of positional certainty based on visual fiducial measurements.

\textcite{russellSwarmRobotForaging} develop a swarm robot foraging algorithm that dynamically deploys and re-deploys sensor motes. 
While the sensors are moved to provide certainty implicitly, they do not model localization quality during their deployment. 

Sensor coverage problems also directly relate to our work. 
In particular, dynamic sensor coverage problems consider moving sensors. 
\textcite{liuDynamicCoverageMobile2013} consider randomly moving mobile sensors and analyze coverage properties. 
Clustering has been applied to dynamic coverage problems~\cite{yuDynamicCoverageControl2022}, but the sensors themselves were considered as mobile units rather than beacons, which are deployed and re-deployed by a moving robot. 
Sensor deployments via actuators are also similar to our problem~\cite{liSensorPlacementSensor2010}. 
However, the problems of deployments of actuators often focus on coverage and do not discuss the coupling of localization and deployment.

In sensor coverage problems, some work incorporate a continuous sensor field intensity~\cite{megerianExposureWirelessSensor2002, wangCoverageProblemsSensor2011}. 
In the attenuated disk model, sensing quality decays with the distance to the sensor. 
This modeling is similar in spirit to our modeling of visual fiducials, but our model focuses as much on directionality as on distance in determining the noise model.

Deployment planning algorithms for heterogeneous robot teams also relate to our work. Such algorithms consider both cases where rewards are known apriori and those in which rewards are randomly distributed~\cite{leeStochasticAssignmentDeploying2021, leeOptimalSequentialStochastic2021, mitchellSequentialStochasticMultiTask2023}. Our problem statement is similar in that there is competition for resources and dependencies imposed by deployment decisions, but we judge the quality of the assignment by the time required to execute the deployment plan rather than as rewards accumulated at each assignment.

\textbf{Localization.} 
Mainstream underwater localization relies acoustic sensors, such as Doppler Velocity Log (DVL), long/short/ultrashort baseline acoustic positioning systems, and multibeam or sidescan sonars~\cite{paull2013auv,petillotUnderwaterRobotsRemotely2019,maurelli2021auv}. While such sensors allow the robot to navigate in large areas, their accuracy depends on a number of external factors, including multipath effect and their overall resolution, making acoustic sensors not best suited to support manipulation tasks. 
Vision-based perception is ubiquitously adopted for many robotics tasks~\cite{cadenaPresentFutureSimultaneous2016}, including underwater~\cite{mcconnellPerceptionUnderwaterRobots2022}, given camera's low cost and ability to capture rich information of the surrounding. The literature classified state estimation methods according to different axes, with one being on whether they minimize reprojection errors of tracked features -- indirect methods, such as ORB-SLAM~\cite{mur-artalORBSLAMVersatileAccurate2015} -- or the alignment error considering image intensity values -- direct methods, e.g., DSO~\cite{engelDirectSparseOdometry2018}. Adding IMUs~\cite{camposORBSLAM3AccurateOpenSource2021} can improve state estimation and including loop closure will allow the odometry estimate to be corrected. Underwater, however, vision-based perception still remains a challenge mainly due to the haze, color loss, and featureless environments~\cite{quattriniliExperimentalComparisonOpen2017,joshiExperimentalComparisonOpen2019}. Given the precision required by the underwater construction task, we rely on visual fiducial markers and extend the operation area of the robot by allowing the robot to move them.

Fisheye cameras are often used to localize mobile robots~\cite{houbenParkMarkingbasedVehicle2019, huangVisionbasedSemanticMapping2018}, but there are no techniques for modeling the quality of features detected using a visual fiducial marker.

Visual fiducial markers have been developed specifically for fisheye cameras~\cite{hajjamiArUcOmniDetectionHighly2020} with the purpose of providing better position information. Other visual fiducial markers have been developed to reduce positioning noise~\cite{benligiraySTagStableFiducial2017}. To our knowledge, no attempt has been made to directly model the uncertainty of detecting visual fiducials. An exploration of the noise properties of visual fiducials is presented by Kalaitzakis et al.~\cite{kalaitzakisFiducialMarkersPose2021}, but the geometry of the noise distribution is unexplored.

\textbf{Free-floating construction systems.} 
We are inspired by the limitations of our autonomous underwater construction system~\cite{lensgrafBuoyancyEnabledAutonomous2023}. This work builds on and extends our autonomous underwater construction robot. Previously, it localized using a single visual fiducial that provided a limited coverage area.

Existing aerial free-floating construction systems commonly make use of fixed-place motion capture systems~\cite{augugliaroBuildingTensileStructures2013, augugliaroFlightAssembledArchitecture2014, hunt3DPrintingFlying2014}. These motion capture systems provide precise, low latency position information but require numerous precisely calibrated cameras with limited coverage area. We want to provide coverage to large areas with limited need for complex fixturing.

\section{Problem Model}\label{sec:model}

We consider the problem of deploying and moving infrastructure dynamically to provide high quality localization information for a set of tasks at known positions in global coordinates. The robot localizes using $m$ beacons which can be placed and moved throughout the mission. The quality of information coming from the beacons depends on how the robot is positioned relative to the beacons. Information from each of the beacons can be combined to increase the localization accuracy. Our goal is to find a mission plan, $\mathcal{A}$, which consists of an ordered set of actions $a_i$. Each $a_i$ can correspond to picking up a beacon, placing a beacon, or completing a task.  

Each of the $n$ tasks, located at $t_i \in \mathbb{R}^3$, requires a high enough precision of localization information. We model the quality of information coming from a beacon $b_i$ using a function $\Sigma(r_i) \mapsto \Sigma_{r_i} \in \mathbb{R}^{3 \times 3}$ which maps relative positions ($r_i$) into covariance matrices that describe the noise distribution of the information coming from the beacon. We assume zero mean error. Information from multiple sensors can be combined by using sensor fusion equations. We use the equations described by \textcite{smith1986representation}. %
Algorithm~\ref{alg:fuse_covariances} shows how we combine noise distributions from multiple sensors.

\begin{algorithm}
\captionsetup{font=small}
\caption{Procedure to fuse covariance matrices of uncertain positions~\cite{smith1986representation}.}
\begin{algorithmic}[1]
    \Require Covariance matrices $\Sigma_1,\dots,\Sigma_n$
    \Ensure Fused covariance matrix $\Sigma$
    \State $\Sigma \gets \Sigma_1$
    \For{$i \in 1\dots,n$}
        \State $K \gets \Sigma (\Sigma + \Sigma_i)^{-1}$
        \State $\Sigma \gets \Sigma  - K \Sigma$
    \EndFor
    \State \Return $\Sigma$
\end{algorithmic}
\label{alg:fuse_covariances}
\end{algorithm}

Each task $t_i$ requires a high enough precision to be completed. We model the precision requirements using a scalar $C_i$ which is obtained by applying a certainty approximation function $C(\Sigma) \mapsto C_i \in \mathbb{R}$ to the fused covariance matrix $\Sigma$. 
We define $C(\Sigma)$ as a function which approximates the probability that location readings are inside of a given error range. Receiving a location reading outside of the error range could cause the robot to fail at its task.

To move a beacon $b_i$, the robot must have precise enough location information for both pickup and placement. This means that beacons must be clustered to provide coverage of one another. For simplicity in our initial exploration, we assume perfect placement of the beacons. 
This assumption is reasonable for moving beacons on error-correcting foundations. In future work, we plan to extend our method to model a decay in the quality of information of moved beacons because of small placement errors.

For simplicity, we also assume a known relative orientation $R$ of the AUV. In practice, the orientation of a free-floating robot can be sensed with a high accuracy using out-of-the-box AHRS boards. An initial calibration step can be used to measure the relative rotation for the set of beacons.

\textbf{1D example.} 
\textbf{}
\begin{figure}[t]
    \centering
    \includegraphics[trim={0 3em 0 3em},clip,width=\linewidth]{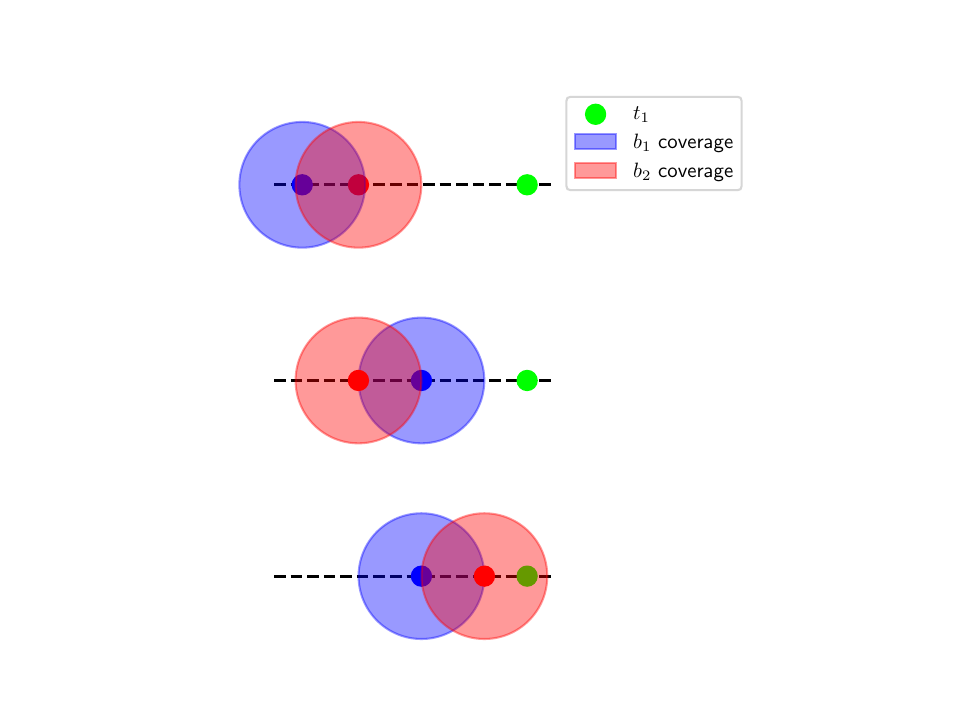}
    \caption{Series of steps to cover a task at $t_1$ with the red and blue beacons.}
    \label{fig:1d-example}
\end{figure}
Consider a robot operating in a 1D world with no collisions. Two beacons, red and blue in Figure~\ref{fig:1d-example}, provide coverage with precision $\Sigma_r(r) = r^2$, that is the quality of information decays quadratically with the distance to the beacon. We set $C(\Sigma) = 1-\Sigma$ because, in 1D, $\Sigma$ is a scalar and can be used directly. The beacons start at positions $b_1 = -0.1$ and $b_2 = 0.1$. The robot is given one task to complete at position $t_1 = .7$. Our fusion function in Algorithm~\ref{alg:fuse_covariances} becomes $\Sigma = r_1^2 - \frac{r_1^2}{r_1^2 + r_2^2}$. The task has a requirement $1 - \Sigma \geq 0.95$. Moving a beacon requires the same certainty. Figure~\ref{fig:1d-example} shows an example of the problem.

We can compute a coverage area for a single beacon and a pair of beacons to guide our creation of a simple mission plan: $[b_i - 0.224, b_i + 0.224]$. To reach and cover our task at position $0.7$, we need to move one beacon to position $.7-0.224 = 0.476$. Moving the beacons will require multiple hops due to their limited coverage area. Our final plan $\mathcal{A}$ is then $\mathcal{A} =$ $\textsc{MoveBeacon}(b_1, 0.324)$, $\textsc{MoveBeacon}(b_2, 0.548)$, $\textsc{complete}(t_1)$. 

In the specific case of assembly a task $t_i$ will represent placing a block at $t_i$'s location. For this application, we write $\textsc{PlaceBlock}(t_i)$ to mean placing a block at location $t_i$. We also replace $\textsc{MoveBeacon}$ with $\textsc{MoveMarker}$ when we are dealing with a reconfigurable visual fiducial marker.

\section{Noise characterization of visual fiducials}

\begin{figure}[t]
     \centering
     \begin{subfigure}[b]{0.21\textwidth}
         \centering
         \includegraphics[trim={0 1em 0 0},clip,height=1.4in]{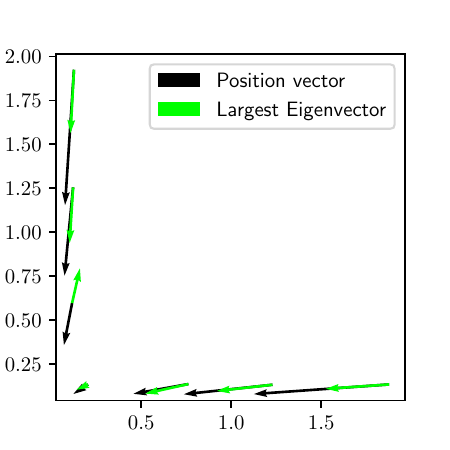}
         \caption{}
     \end{subfigure}
     \begin{subfigure}[b]{0.25\textwidth}
         \centering
         \includegraphics[trim={0 2em 0 0},clip,height=1.4in]{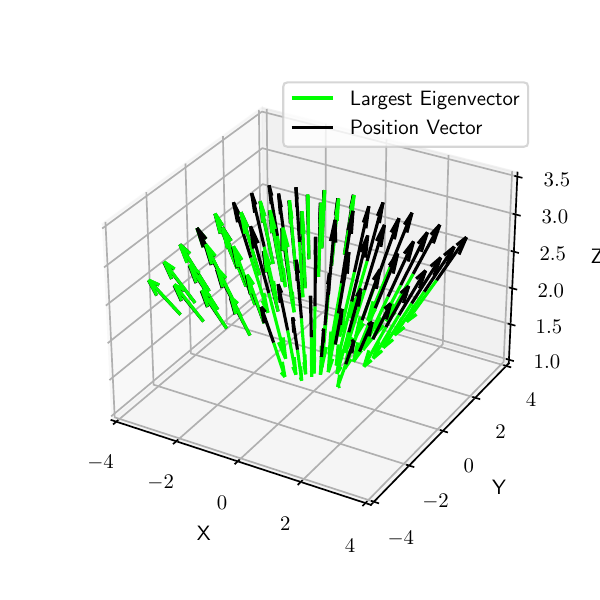}
         \caption{}
     \end{subfigure}
    \caption{(a) Predicted and measured largest eigenvector directions in real world experiment. The arrows extend from the marker's position. (b) Results from simulated corner noise. In both cases the predicted and measured directions closely match.}
    \label{fig:eigenvector-directions}
\end{figure}

The first step to implementing our assembly planning and localization method is to accurately model the noise distribution of visual fiducial markers. To understand how the noise distribution varies based on the relative position between the fisheye camera and a marker, we built a simulator. The simulator applies Gaussian distributed noise to the distorted corner positions of the visual fiducial, then undistorts them using the Kannala-Brandt
Camera model~\cite{kannala2006generic} -- typically used for fisheye lenses --  and solves the Perspective-n-point problem. This simulation captures the important sources of noise: sensor noise and barrel distortion. We experimentally validate our simulator and noise model in \Cref{sec:validation-of-noise}.

Figure~\ref{fig:eigenvector-directions} shows outputs from a real world experiment (a) and from our simulation (b). 
We found that the noise distribution was highly structured and could be predicted using only two values: the largest eigenvector and eigenvalue. 
Further, the eigenvalue is parallel to the position vector. 
In the remainder of this section, we discuss how we predict the two components.

\subsection{Scale noise}

\begin{figure}[t]
    \centering
    \includegraphics[width=\linewidth]{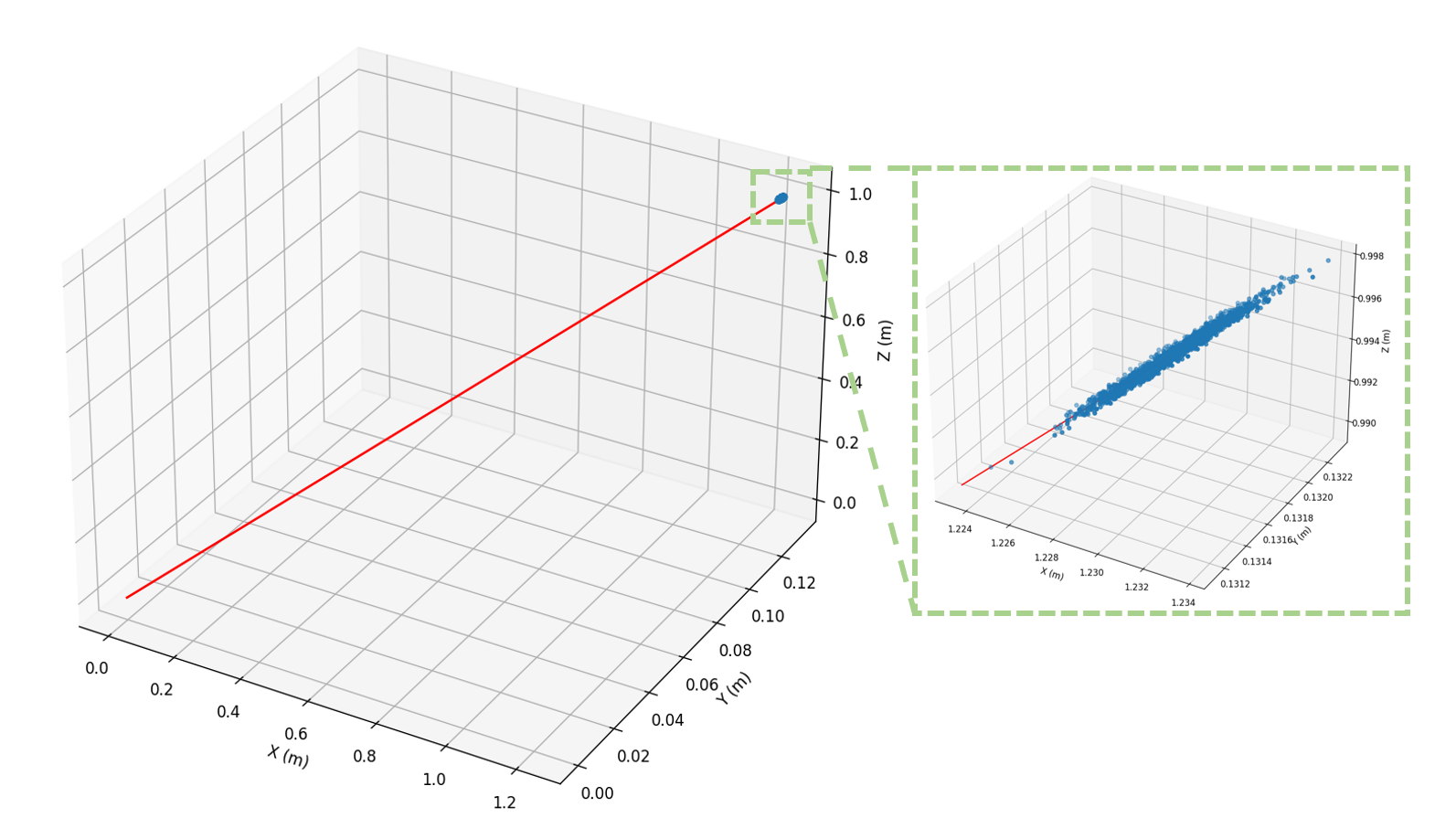}
    \caption{Position readings from a static fisheye lens camera for a static visual fiducial marker.}
    \label{fig:fiducial-noise}
\end{figure}

In our experiments on fiducial marker noise when viewed through a fisheye lens, we found that the axis of largest noise very consistently pointed towards the camera. 
Figure~\ref{fig:fiducial-noise} shows an example of a set of relative position measurements for a single visual fiducial measured by a static camera. 
The position vector is marked in black and aligns closely with the largest eigenvector of the noise distribution (green). 
We noticed this phenomenon to be consistent across various positions. 
The largest eigenvector dominates the noise distribution, and the other two are an order of magnitude smaller. 
We provide empirical evidence for this observation in Section~\ref{sec:validation-of-noise}.

\subsection{\texorpdfstring{A definition of $\Sigma(p)$}{A definition of Σ}}

\begin{algorithm}[b]
\captionsetup{font=small}
\caption{Find covariance matrix $\Sigma_p$ for a relative position $p$.}
\label{alg:cov-matrix-prediction}
\begin{algorithmic}[1]
\Require $p$, $\beta$, $\lambda_i$
\Ensure $\Sigma_p$

\State $\lambda_* \gets \beta(p)$
\State $v_1 \gets \frac{p}{\| p \|}$
\State $v_2 \gets \textsc{orthogonal}(p)$
\State $v_3 \gets \textsc{orthogonal}(p, v_2)$

\State $\Lambda \gets \textsc{hstack}(v_1, v_2, v_3)$
\State $S \gets \textsc{diag}(\lambda_*, \lambda_i, \lambda_i)$

\State \Return $\Lambda S \Lambda^{-1}$

\end{algorithmic}
\end{algorithm}

Algorithm~\ref{alg:cov-matrix-prediction} shows our covariance matrix prediction procedure. It accepts as arguments $p$, the relative position, $\beta$, a predictor of the largest eigenvalue and $\lambda_i$, an upper bound on the two smaller eigenvalues.

The largest eigenvalue, $\lambda_*$, is predicted using a spline which is calibrated on experimental data. $v_1$, $v_2$, and $v_3$ are our predictions of the eigenvectors. The largest eigenvector, $v_1$, is predicted to be the normalized position vector. The other two eigenvectors, $v_1$, and $v_2$ are predicted to be orthogonal to $p$ and one another. Line 6 accumulates the eigenvalues into a $3 \times 3$ diagonal matrix. The matrices $\Lambda$ and $S$ are multiplied together to produce a predicted covariance matrix which has the predicted eigenvalues and eigenvectors.

\section{Finding feasible assembly plans in practice}

We model the assembly process as a set of manipulation tasks located at points in 3D space. 
The probability of completing a task $C(\Sigma_r)$ is the probability that a block dropped will be inside of the acceptance area of the slot it is aimed at. 
As a structure is built, it can occlude the view of markers, therefore, the markers must be continually moved as the structure is erected to increase $C(\Sigma_r)$.

\subsection{\texorpdfstring{Computing $C(\Sigma_r)$}{Defining C(Σ\textunderscore r)}}

To enable planning during the construction process, we define $C:\Sigma_r \mapsto \mathcal{R}$, which takes as input a covariance matrix and outputs the probability of successfully dropping a block. 
A block or marker is successfully placed if the robot decides to initiate the placement action within an acceptable range of the ideal position. 
This acceptable range ($\alpha$) is dictated by the design of the error-correcting construction foundation. 
Assuming that the uncertainty in position is well approximated with a multivariate Gaussian, $C(\Sigma_r)$ is the probability that a random sample position drawn from $\mathcal{N}(0, \Sigma_r)$ is within the sphere of radius $\alpha$. 

Analytically computing this probability is challenging. 
Instead, we use a conservative approximation. 
The largest eigenvalue ($\lambda_*$) is a conservative estimate of the standard deviation of noise in any direction. 
So, we assume that the noise distribution along all three coordinate axes is independent and is equal to $\lambda_*$. This approximation results in the closed form estimate $C^*(\Sigma_r)$:

\begin{equation}
\label{eqn:cr_metric}
C(\Sigma_r) \ge \erf{\left( \frac{\alpha}{\sqrt{\lambda_*} \sqrt{2}} \right)}^3 = C^*(\Sigma_r). 
\end{equation}

\subsection{A layer-based approach for assembly}

To assemble a structure, the robot must manipulate markers which can rest on top of blocks in order to localize while blocks are placed. When placed, the blocks can obscure sight of the markers. Planning around obscured markers introduces difficult nonlinearities into the constraints for any solver hoping to find feasible solutions.

To construct feasible plans efficiently without needing to model occlusions between the structure as it is erected and the markers, we propose a layer-by-layer algorithm. Algorithm~\ref{alg:layer-based-traversal} shows our strategy for generating feasible plans for a structure with $n$ slots, using $m$ markers.

Our algorithm works by dividing the blocks into layers $l_1,\dots,l_h$, where $l_1$ is the bottom layer of blocks, $l_2$ is the layer to be placed above $l_1$, and so on with $l_h$ being the topmost layer of blocks. For each $i \in \{1,2,\dots,h\}$, we first cluster $l_i$ to obtain clusters of width at most $r$, where $r$ is an empirical determination of a marker's coverage radius based on the bound in Equation~\ref{eqn:cr_metric}. In our implementation, this is achieved using a subroutine \textsc{ClusterUntilRadius}$(l_i,r)$. This subroutine performs $k$-means clustering on $l_i$ using a value of $k$ that is tuned, via binary search, to be the minimum possible such that all cluster widths are at most $r$.

The sub-procedure \textsc{ExtractCenters} finds the center point of each cluster. The cluster centers are then passed into \textsc{FindTour} which computes a tour of the cluster centers $O$ which has elements that index the clusters.

In each cluster produced by \textsc{ClusterUntilRadius}$(l_i,r)$, we select $m$ points to serve as marker destinations. We choose the marker destinations to be the $m$ farthest points from each other in the cluster. We then use the sub-procedure \textsc{WalkToCoverage} to transition the markers between destinations. We achieve this via a simple hopping strategy, like the one discussed in \Cref{sec:model}. %
This strategy repeatedly hops one marker to the outside of the other marker's coverage area, resembling a ``gait'' if one imagines the markers to be a robot's feet. In this way, we position the markers at the $m$ destinations within a cluster, place the blocks within that cluster, and repeat for successive clusters in that layer. Since the maximum cluster width is the coverage radius of a marker, we can ensure that each marker is always covered by another marker, enabling us to continue the gait after the blocks in a cluster have been placed.

After blocks have been placed around each of the markers in the cluster, the markers must be moved to make room for the remaining blocks. To do this, the loop in \crefrange{alg:ln:layer-by-layer:last-markers:start}{alg:ln:layer-by-layer:last-markers:end} iterates over each marker, moves it on top of the nearest block to it, and then places a block where the marker used to be. If the cluster has fewer than $m$ points, two markers might share the same nearest neighbor and cause the markers to be placed in the same location. We avoid this edge case by requiring a minimum cluster size of $m$ in \textsc{ClusterUntilRadius}$(l_i,r)$; this is feasible via a reasonable assumption that $m$ markers can be placed within radius $r$.

\begin{algorithm}[tb]
\captionsetup{font=small}
\caption{Layer-by-layer traversal.}
\begin{algorithmic}[1]
    \Require Structure with slots $S = \{s_1,\dots,s_n\}$, marker positions $M = \{b_1,b_2,\dots,b_m\}$
    \Ensure Assembly plan $\mathcal{A}$
    \State Divide $S$ into layers $l_1, \dots, l_h$
    \State $\mathcal{A} \gets [ ]$
    \For{$i \in \{1,\dots,h\}$}
        \State $\mathcal{C} \gets \textsc{ClusterUntilRadius}(l_i,r)$
        \State $C \gets \textsc{ExtractCenters}(\mathcal{C})$
        \State $c_1,c_2,\dots,c_k \gets \textsc{FindTour}(C)$
        \Comment{efficient tour on cluster centers}
        \For{$j=1$ to $k$} \Comment{process clusters in tour order}
            \State $C_j \gets \textsc{ClusterOf}(c_j)$
            \State $\mathcal{A} \gets \mathcal{A}$.\textit{extend}\big(\textsc{WalkToCoverage}$(M,c_j)$\big) \Comment{move markers into cluster}
            \For{$s \in C_j \setminus M$} \Comment{place blocks in slots unoccupied by markers}
                \State $\mathcal{A} \gets \mathcal{A}.\textit{append} \big(\textsc{PlaceBlock}(s)\big)$ 
            \EndFor
            \For{$s \in M$} \Comment{move markers to place remaining blocks}\label{alg:ln:layer-by-layer:last-markers:start}
                \State $p \gets \textsc{NearestNeighbor}(s, C_j)$
                \State $p \gets p + (0,0,1)$
                \State $\mathcal{A} \gets \mathcal{A}.\textit{append} \big(\textsc{MoveMarker}(s, p)\big)$
                \State $\mathcal{A} \gets \mathcal{A}.\textit{append} \big(\textsc{PlaceBlock}(s)\big)$\label{alg:ln:layer-by-layer:last-markers:end}
            \EndFor
        \EndFor
    \EndFor
    \State \Return $\mathcal{A}$
\end{algorithmic}
\label{alg:layer-based-traversal}
\end{algorithm}

\section{Experiments}
\label{sec:experiments}

We implement our reconfigurable visual fiducial localization system in both hardware and simulation, and the results are described in the following.

\subsection{Experimental setup}

 Our hardware implementation is deployed on the Droplet AUV system~\cite{lensgrafBuoyancyEnabledAutonomous2023, lensgrafDropletAutonomousUnderwater2021}. A ROS node holds the known global position of the visual fiducial markers. As the markers are moved, the node is notified and then marker readings are offset accordingly before performing the sensor fusion steps described in Algorithms~\ref{alg:fuse_covariances} and~\ref{alg:cov-matrix-prediction}.

For viewing visual fiducial markers, the robot is equipped with FLIR Blackfly camera with a Senko fisheye lens mounted facing downards. The camera is mounted in a 3 inch Blue Robotics acrylic enclosure with a dome. Figures~\ref{fig:two-hop-success},~\ref{fig:hero-image} show the camera mounted on the robot and Figure~\ref{fig:reconfigurable-visual-fiducials} (b) shows how the scene looks through the camera.

\subsection{Hardware testing}
\label{sec:hardware-experiments}

\begin{figure}[t]
     \centering
     \begin{subfigure}[b]{0.21\textwidth}
         \centering
         \includegraphics[height=1.2in]{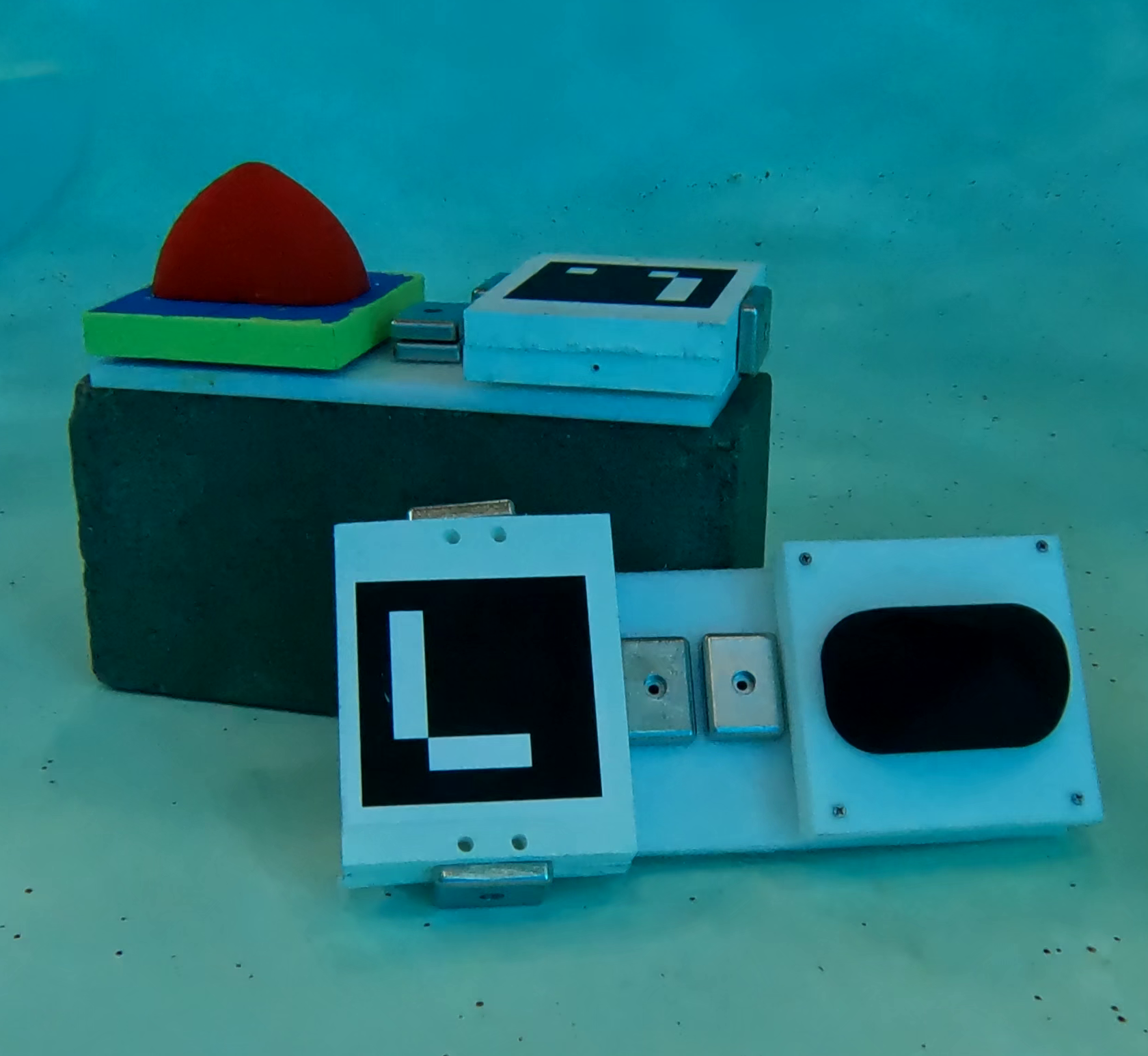}
         \caption{}
     \end{subfigure}
     \begin{subfigure}[b]{0.25\textwidth}
         \centering
         \includegraphics[height=1.2in]{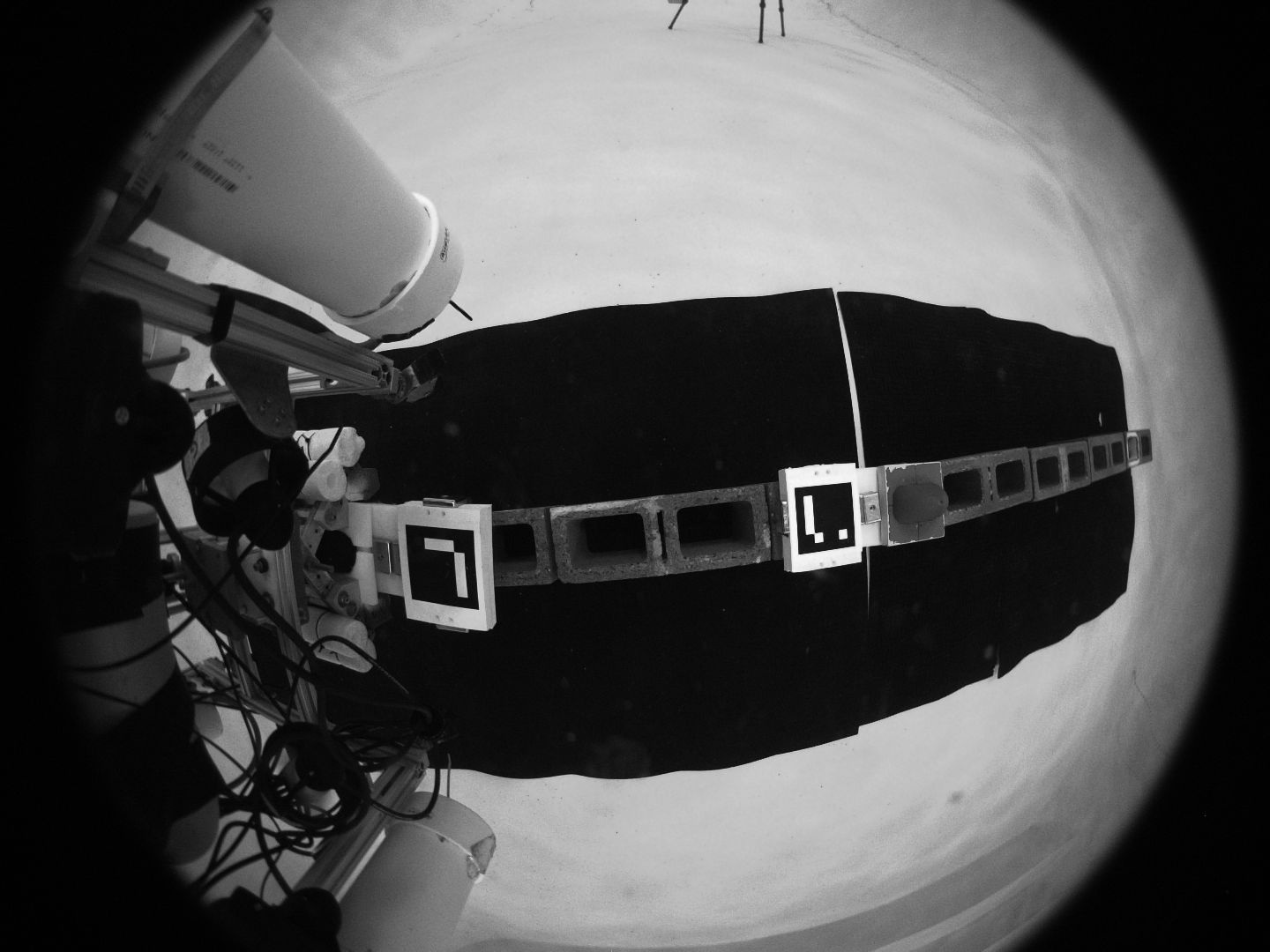}
         \caption{}
     \end{subfigure}
    \caption{(a) Reconfigurable visual fiducial design. (b) Robot's eye view during assembly with the reconfigurable fiducials.}
    \label{fig:reconfigurable-visual-fiducials}
\end{figure}

\begin{figure}[t]
    \centering
    \includegraphics[width=\linewidth]{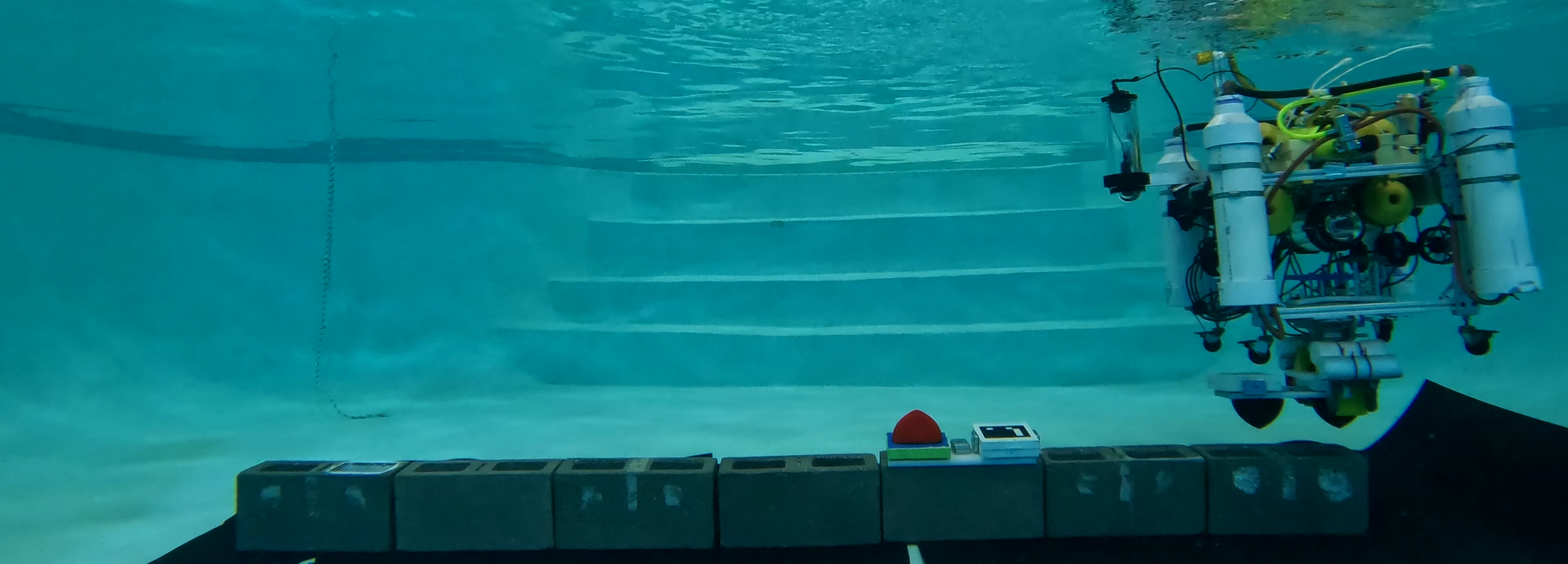}
    \caption{The AUV performing the last placement of the two hop maneuver.}
    \label{fig:two-hop-success}
\end{figure}

To validate the concept of reconfigurable visual fiducials in practice, we mounted visual fiducial markers on our error correcting connector geometry ~\cite{lensgrafBuoyancyEnabledAutonomous2023}. 
We show the design in Figure~\ref{fig:reconfigurable-visual-fiducials}. 
The fiducials provide error correction during both pickup via the top handle (red in Figure~\ref{fig:reconfigurable-visual-fiducials}) and placement.

We tested the concept of walking behaviors by implementing a 1D hopping strategy. 
The robot was able to successfully perform manipulation tasks in a large area with a side of 2.8 meters using cement blocks placed by hand. 
This side is 57\% longer than the widest side usable by our previous implementation.
Note that the side of the line covered was bounded by the physical width of the pool and not the robot's ability to stack blocks successfully. 
Figure~\ref{fig:two-hop-success} shows the AUV completing the maneuver.

To verify that the visual fiducial marker fusion procedure reduces the localization noise, we recorded the measured positions and the number of valid marker readings when the robot was still. 
When only one marker was determined as valid, the standard deviation of position measurements on the X axis was \SI{4.3}{cm}, but when two markers were valid, the standard deviation was reduced to \SI{0.69}{cm}; around $6\times$ improvement. On the Y axis, we observe similar improvements: from \SI{6.0}{cm} to \SI{1.5}{cm}; around $4\times$ improvement.

\subsection{Validation of noise model}

\label{sec:validation-of-noise}

To validate our prediction model of covariance matrices given in Algorithm~\ref{alg:cov-matrix-prediction}, we conducted both real world and simulation experiments. Our simulator projects a marker at a given relative position into the camera and offsets the corners according to a Gaussian distribution. This noise addition simulates the effects of pixel flicker on corner detection.

Figure~\ref{fig:eigenvector-directions} (a) shows the results of our physical testing. We arrayed visual fiducials in an area spanning about 1.4 meters on the positive X and Y axes. This setup mimics the distances between markers used in practice. We predict the direction of the largest eigenvector as the median of the measured relative positions. We found that the predicted direction is accurate to within $4$ degrees. This result shows that computing the direction of the largest eigenvector as the position vector is effective in practice.

To more extensively test our prediction algorithm, we conducted a test using a spline predictor of the largest eigenvalue trained on 6250 noise distributions generated at known relative positions in our simulator. We set the upper bound for the smallest two eigenvalues to $10^{-4}$. We set $\alpha$ to \SI{2}{cm} for Equation~\ref{eqn:cr_metric}. We tested the trained predictions from Algorithm~\ref{alg:cov-matrix-prediction} using 5184 test relative positions not present in the training data. 

To evaluate the quality of Algorithm~\ref{alg:cov-matrix-prediction}, we checked whether its predictions were more or less conservative than measured covariance matrices in the test dataset. We evaluated whether Algorithm~\ref{alg:cov-matrix-prediction} produced more or less conservative results by comparing the bound in Equation~\ref{eqn:cr_metric} for the predicted covariance matrix $\Sigma_r^*$ against the measured covariance matrix $\Sigma_r$. We found that in 98.3\% of cases $C^*(\Sigma_r^*) \leq C^*(\Sigma_r)$. In the other 1.7\% of cases, $C^*(\Sigma_r)$ was only smaller than $C^*(\Sigma_r^*)$ by at most 3.4\%. To avoid this case, we configure our planning algorithm so two markers are always visible.

When combining the two worst predictions using Algorithm~\ref{alg:fuse_covariances}, we find that the fused covariance matrix is a conservative estimate. We combined the covariance matrices of the two worst over-predictions of $C^*(\Sigma_r)$ for both measured and predicted covariances. We found that the predicted $C^*(\Sigma_r)$ is 40\% lower (59\%) than the measured bound (99\%). 

Our estimate of the noise is often very conservative, but we will show in the following section that large structures still have a high predicted success probability when planned using Algorithm~\ref{alg:layer-based-traversal}.

\subsection{Assembly algorithm}

\begin{figure}[t]
     \centering
     \begin{subfigure}[b]{0.25\textwidth}
         \centering
         \includegraphics[width=\textwidth]{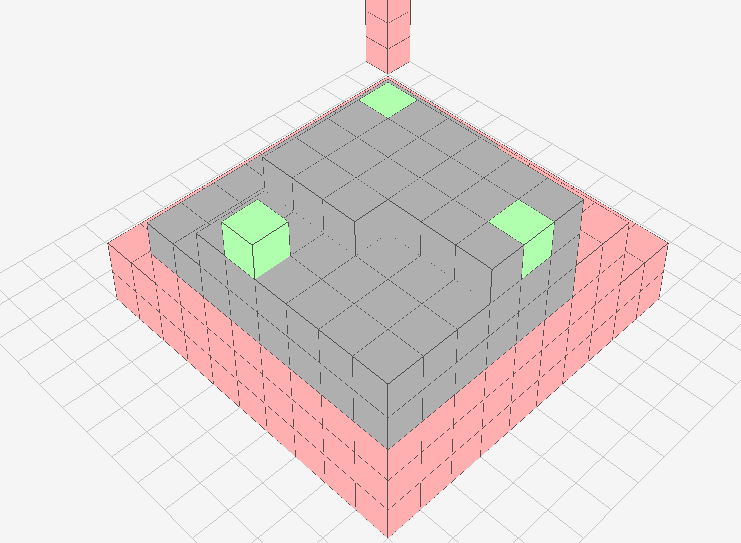}
         \caption{}
     \end{subfigure}
     \begin{subfigure}[b]{0.20\textwidth}
         \centering
         \includegraphics[width=\textwidth]{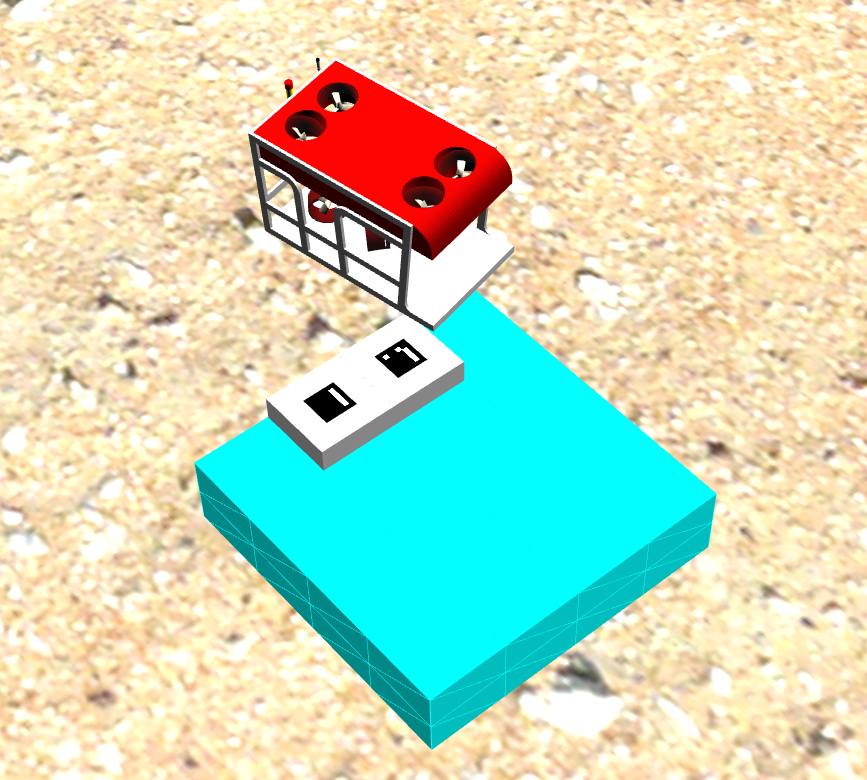}
         \caption{}
     \end{subfigure}
    \caption{(a) Visualization of plan checker. Green blocks represent markers. (b) Robot moving markers in the DAVE simulator.}
    \label{fig:sim-testing}
\end{figure}

To test Algorithm~\ref{alg:layer-based-traversal} we use both a hand crafted plan checker and a full underwater robot 3D simulator called DAVE~\cite{zhangDaveSimulator}. Our plan checker checks lines of sight to visual fiducials and computes the predicted certainty of every step. We use the underwater robot simulator to determine whether a plan which is feasible according to the plan check is also feasible under realistic control noise. We ran a construction process with 74 build steps. Figure~\ref{fig:sim-testing} (b) shows the partially completed structure in the DAVE simulator. With realistic control noise, the robot kept at least two markers in view 100\% of the time.

\begin{table}[tb]
\centering
\caption{Effect of increasing the cluster radius on plan efficiency and probability of success. After a certain cluster size, occlusions make construction impossible.}
\label{table:r-change-effects}
\begin{tabular}{|c | c | c |} 
 \hline
 $r$ & Predicted $P(\text{success})$ & \# steps \\
 \hline
 2.5 & 0 & 226 \\
 2.0 & 0.82 & 240 \\
 1.5 & 0.96 & 259 \\
 1.0 & 0.99 & 345 \\
 \hline
\end{tabular}
\end{table}

As the number of manipulation steps increases, the time to build a structure increases. Our layer-based construction (Algorithm~\ref{alg:layer-based-traversal}) takes an input a parameter $r$ which describes the radius of clusters used for marker placement. Between each cluster, the markers are moved in an expensive walking procedure, so increasing $r$ could 
improve the efficiency of the construction process but at the cost of reducing the reliability. Table~\ref{table:r-change-effects} shows how changing $r$ affects the number of steps required to build a 200 block structure and the certainty afforded during construction with three markers. We measure the predicted $P(\text{success})$ as the product of the probability of success of every state in the construction process. We also tested a pyramidal structure containing 1800 blocks. The assembly planning algorithm took about 5 minutes to plan the structure and the predicted probability of success for $r = 1.5$ is $91\%$.

\section{Conclusions \& future work}

This paper proposes a novel strategy for localizing relative to error correcting structures while planning the construction process. Our method is shown to work in practice at small scale and at large scale in simulation. We also show the robot being able to reliably complete the ``hopping'' strategy for moving the markers, extending the area for assembly.

We plan to improve the scale and quality of our hardware implementation. Planning for large scale construction with heterogeneous materials will require adaptive and flexible clustering strategies. Materials which are not well described by a bounding box may require more sophisticated strategies for avoiding occlusions.

Our assembly process is currently limited to structures which have only a single connected component per layer. If there is more than one connected component in a layer, the markers can become stranded as they are lifted up the structure. In the future, we plan to explore ways to increase the flexibility of our method.

\printbibliography

\end{document}